\documentclass[runningheads]{llncs}

\usepackage[ruled,vlined]{algorithm2e}
\SetKwComment{tcp}{\color{blue}\#\ }{}
\usepackage{soul}

\usepackage{tabularx}

\usepackage{graphicx}
\usepackage{booktabs}
\usepackage{makecell}

\usepackage[accsupp]{axessibility}  % Improves PDF readability for those with disabilities.

\usepackage[T1]{fontenc}

\usepackage{graphicx}

\usepackage[breakable]{tcolorbox}
  {\begin{tcolorbox}[colback=yellow,
                     colframe=white,
                     boxsep=0pt,
                     left=0pt,
                     right=0pt,
                     top=0pt,
                     bottom=0pt,
                     center, 
                     valign=top, 
                     before skip=0pt, 
                     after skip=0pt,
                     width=\textwidth,
                     breakable,
                     boxrule=0pt,
                     frame empty,
                     sharp corners
                 ]
             \setlength{\parindent}{\defaultparindent}
             }%
  {\end{tcolorbox}}
  
\begin{document}
\title{LoGDesc: Local geometric features aggregation for robust point cloud registration} 
\titlerunning{LoGDesc}
\author{Karim Slimani\inst{1}\and
Brahim Tamadazte\inst{1} \and
Catherine Achard\inst{1}}
\authorrunning{K. Slimani et al.}
\institute{ISIR, Sorbonne Université, CNRS UMR 7222, INSERM U1150,
        4 place Jussieu, Paris, France \\
Corresponding Author: \email{karim.slimani@isir.upmc.fr}\\
% \email{brahim.tamadazte@cnrs.fr}\\
% \email{catherine.achard@sorbonne-universite.fr}\\
}
\maketitle              % typeset the header of the contribution

\begin{abstract}
This paper introduces a new hybrid descriptor for 3D point matching and point cloud registration, combining local geometrical properties and learning-based feature propagation for each point's neighborhood structure description. The proposed architecture first extracts prior geometrical information by computing each point's planarity, anisotropy, and omnivariance using a Principal Components Analysis (PCA). This prior information is completed by a descriptor based on the normal vectors estimated thanks to constructing a neighborhood based on triangles. The final geometrical descriptor is propagated between the points using local graph convolutions and attention mechanisms. The new feature extractor is evaluated on \textit{ModelNet40}, \textit{Bunny Stanford} dataset, \textit{KITTI}, and MVP (Multi-View Partial)-RG for point cloud registration and shows interesting results, particularly on noisy and low overlapping point clouds.\\
\textit{The code will be released after publication.}

\end{abstract}

\section{Introduction}
\label{sec.intro}
The recent development of new generations of three-dimensional sensors, such as LIDAR and Kinect, offers many applications in computer vision, robotics, 3D printing, graphics, etc. Unlike traditional visual sensors (cameras), these sensors provide a 3D representation of the observed scene in the form of a 3D point set called a point cloud, providing important cues for analyzing objects and environments~\cite{schwarz2010mapping}. These technological advances have enabled us to overcome the recurrent problem of directly estimating scene depth that characterizes traditional visual sensors. The field of view of 3D sensors is relatively limited in many applications, such as robotic navigation, autonomous vehicles, and manipulation tasks. As a result, several scientific questions have emerged relating to 3D reconstruction~\cite{wang2019applications}, mapping~\cite{haala2008mobile}, object or scene recognition~\cite{shi2020points}, pose estimation~\cite{aldoma2012tutorial}, medical imaging~\cite{audette2000algorithmic}, etc. Nevertheless, \textit{3D point registration}, also known as scan matching or point cloud alignment, is undoubtedly the problem that has received the most interest from the robotics, graphics, and computer vision communities~\cite{pomerleau2015review}. A typical registration problem involves aligning two or more point clouds in a three-dimensional coordinate system acquired from different viewpoints into a unified coordinate system. This means finding the optimal spatial transformation (rotation, translation, and potentially scale) that aligns the given point clouds. The registration problem can be tackled from the optimization angle, opening the way to so-called iterative methods. Most of the existing registration methods are formulated by minimizing a geometric projection error through two processes: correspondence searching and transformation estimation, which are repeated until convergence is reached; this is the case in the two most used registration methods, ICP (Iterative Closest Point)~\cite{besl1992method} and RANSAC for instance. For a long time, these methods were the benchmark in point cloud registration, thanks to their efficiency and simplicity of implementation. However, their performances can deteriorate under unfavorable conditions, often leading them to converge on a local minimum due to non-convexity.

To overcome these limitations, features-based methods attempt to extract locally significant and robust geometric descriptors. These descriptors can be either local (extracted from the interesting part of the point cloud)~\cite{yang2016fast}, global (generated by encoding the geometric information of the whole point cloud)~\cite{zhou2016fast} or hybrid, combining local and global descriptors)~\cite{avidar2017local}. Each descriptor has advantages and disadvantages, depending on the point clouds to be aligned, but none is precise, robust, and versatile. 

%
% --------------------
\section{Related Work and Contributions}
% --------------------
%
The emergence of deep learning across multiple disciplines has likewise enhanced 3D registration techniques in several aspects. Among the contributions of deep learning, the development of learned descriptors has brought significant progress in point cloud registration~\cite{dong2020registration}. The feature learning approach, PRNET~\cite{wang2019prnet}, is designed to extract features invariant to rigid transformations, addressing the unordered nature of point clouds and leveraging local surface characteristics. Similarly, R-PointHop~\cite{r_pointhop} extracts the descriptors by utilizing various neighborhood sizes facilitated by a Local Reference Frame (LRF) established for each point through its nearest neighbors. {To achieve rotation invariance, 3DSmoothNet}~\cite{gojcic2019perfect} {projects a voxelized smoothed density value representation to the LRF before feeding it to 3D CNNs.} These approaches have demonstrated promising outcomes on geometric registration benchmarks. However, transforming points to voxels can incur substantial computational expenses, particularly with dense point clouds. Besides, PointNet~\cite{qi2016pointnet} processes the point cloud by mapping it into a learned canonical space and employing a symmetric function to ensure the output remains unaffected by the input points' order. The popular DGCNN~\cite{dgcnn} method enhances local property captured between points by implementing successive convolutional layers. It constructs input graphs based on each point's $K$ Nearest Neighbors that are dynamically refreshed at each layer. 

Likewise, learning hybrid descriptors (combining local and global information) is possible. Recent approaches incorporate local and global features into a transformer module, as in DCP~\cite{dcp}. These models enhance traditional transformers by integrating additional geometric information, including pairwise distances and triplet angles, as exemplified by GeoTransformer~\cite{geotransformer}.

RoCNet++~\cite{slimani2024rocnet++} investigated a new descriptor that encodes the local geometric properties of the surface, \textit{i.e.}, each point is characterized by all the triangles formed by itself and its nearest neighbors. Casspr~\cite{xia2023casspr}{investigated cross-attention transformers fusing point-wise and sparse voxel features to capture information at low and fine resolutions.} Soe-net~\cite{xia2021soe}{ introduced the \textit{PointOE} module to capture local structures by analyzing patterns from multiple spatial orientations}. 

In this paper, we present a new hybrid descriptor \emph{LoGDesc} for point cloud registration, exploiting geometric properties given the very local structure of the points and enhancing their robustness to noise by feeding them to learning modules. The first main contribution is using the normals to the planes containing the triangles formed by each point and its nearest neighbors to estimate a single normal vector on each point by weighting each normal's support by the triangle area's function. Secondly, a local PCA (Principal Component Analysis) allows computing a Local Reference Frame (LRF), the anisotropy ($A$), the omnivariance ($O$), and the planarity ($P$) for each point. The last three functions complete the 3D coordinates to make a robust first vector of features. Then, the previously estimated normals are projected in the LRFs to ensure rotation-invariant descriptors. Finally, the information in each point descriptor is propagated locally to globally thanks to $KNN$-based graphs followed by a \textit{self-attention} mechanism. 

%
% -------------------
\section{Problem Statement and Proposed Method} \label{sec.method}
% -------------------
%
% ---------------
\subsection{Problem Statement}
% ---------------
%
Let us define two point clouds $\boldsymbol{X}$ and $\boldsymbol{Y}$ such that 
$\boldsymbol{X}=\{ \boldsymbol{x}_1,...,\boldsymbol{x}_i,...,\boldsymbol{x}_M\} \subset \mathbb{R}^{3\times M}$ and $ \boldsymbol{Y}=\{ \boldsymbol{y}_1,...,\boldsymbol{y}_j,...,\boldsymbol{y}_N\} \subset \mathbb{R}^{3\times N}$, where each $\boldsymbol{x}_i$ and $\boldsymbol{y}_j$ are the 3D coordinates of the $i^{th}$ and $j^{th}$ points in the source and destination point clouds, respectively. Suppose the two point clouds are at least partially overlapping, \textit{i.e.}, there are $C$ pairs of matches between $\boldsymbol{X}$ and $\boldsymbol{Y}$, forming two sets of matches $\boldsymbol{\bar{X}} \subset \mathbb{R}^{3\times C} $ and $\boldsymbol{\bar{Y}} \subset \mathbb{R}^{3\times C} $, where $C \leq min(M,N)$.

The purpose of point cloud registration is to estimate the rigid transformation matrix $\mathbf{T}_{XY} = 
\begin{bmatrix}
   \mathbf{R}_{xy} & \mathbf{t}_{xy} \\
   0_{1\times3}& 1 \\
\end{bmatrix}$ which minimizes the Euclidean distance between $\boldsymbol{\bar{X}}$ and the transformed $\boldsymbol{\bar{Y}}$: 
\begin{equation}\label{eq.transrigid}
\mathbf{R}_{xy}, \mathbf{t}_{xy} = \underset{\mathbf{R^*}_{xy}, \mathbf{t^*}_{xy}}{argmin}\bigg(~\mathrm{d}\big(\boldsymbol{\bar{X}},\mathbf{R^*}_{XY}\boldsymbol{\bar{Y}} + \mathbf{t^*}_{xy}  \big) \bigg) 
\end{equation}
with $\mathbf{R^*}_{xy} \in \mathit{SO}(3) $ being all admissible  $3 \times 3$ matrices, $\mathbf{t^*}_{xy} \in \mathbb{R}^3$ all admissible 3D vectors and $\mathrm{d}$ the Euclidean distance between the two sets of paired points which can be formulated by: 
$~\mathrm{d}\big(\boldsymbol{\bar{X}},\mathbf{R^*}_{XY}\boldsymbol{\bar{Y}} + \mathbf{t^*}_{xy}  \big)= \sum^C_{c=1} \| (x_c - \mathbf{R^*}_{xy} \cdot y_c +\mathbf{t^*}_{xy}) \|_2$.

%
% ----------------
% \subsection{Previous Work: RoCNet}\label{sec.previous_work}
\subsection{Registration overall architecture}\label{sec.previous_work}
\begin{figure}[!h]
\centerline{\includegraphics[page=1, width=.8\columnwidth]{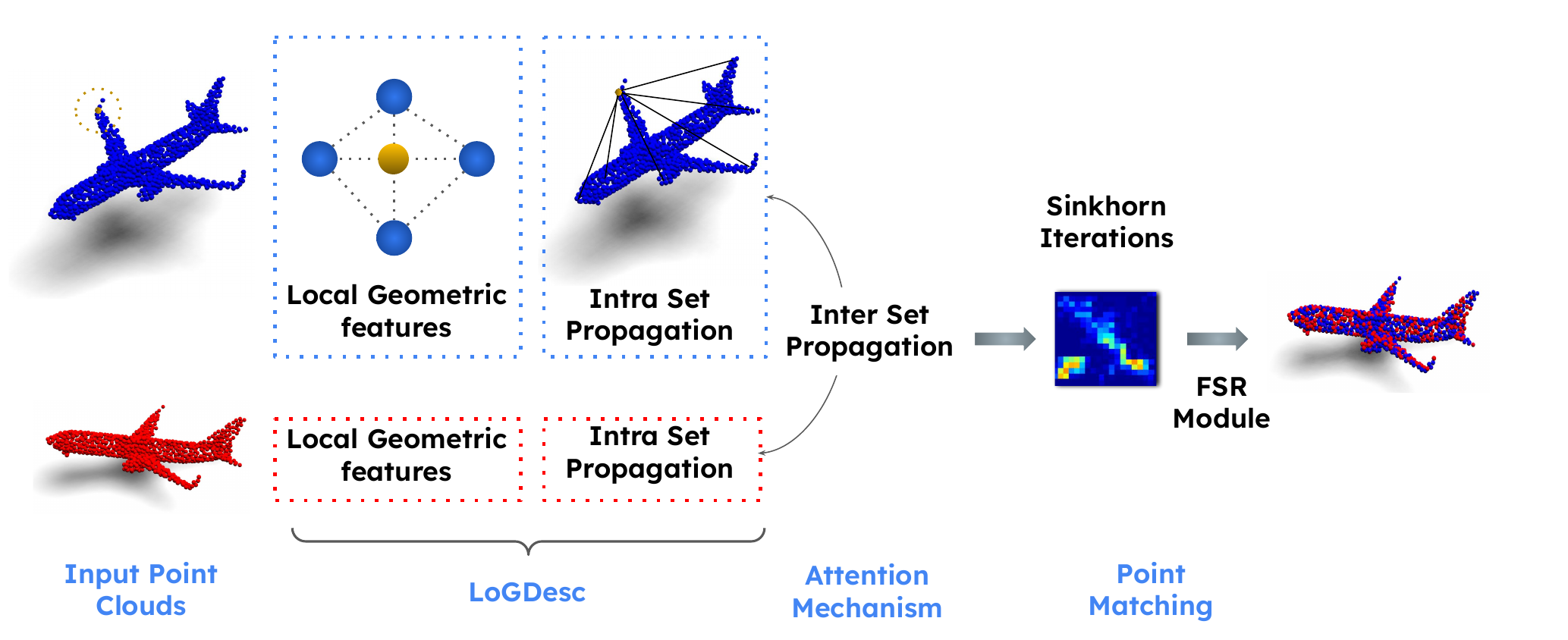}}
\caption{Summary of the point cloud registration method.}
\label{fig.abstr}
\end{figure}

%
% ----------------
%
The hybrid feature extractor proposed in this paper is assessed in the point cloud registration challenge by integrating it in an architecture inspired by the state-of-the-art methods:~\cite{slimani2023rocnet,slimani2024rocnet++,mdgat}, as depicted in Figure~\ref{fig.abstr}. The overall architecture is described in the following subsections.  

%
% ---------------
\subsubsection{Feature extraction}\label{subsec.descriptor}
The first step of the algorithm is to extract the feature vectors $\boldsymbol{f}_i \in \mathbb{R}^{d}$ and $\boldsymbol{h}_i \in \mathbb{R}^{d}$ for each source and target point, respectively. This is a crucial step in learning point cloud registration methods since the accuracy of the estimated rigid transformation explicitly depends on the accuracy of the matching process, which in turn depends on the similarity between the corresponding pairs of points from the two point clouds. To this end, we propose a new descriptor called  \emph{LoGDesc}. The main principle of  \emph{LoGDesc} is to use local geometric properties to capture the patterns of the local structure of each point and to learn to propagate this local information globally within each point cloud using attention mechanisms. The section~\ref{sec.logdesc} details the proposed feature extraction and aggregation.
%
% ----------------
\subsubsection{Normal encoder attention mechanism}\label{subsec.transformer}
Once the features are extracted using  \emph{LoGDesc}, we propose propagating their information within each point cloud and between the two point clouds using a geometric transformer. This allows the algorithm to learn an inter-sets contextual understanding of these features. To do so, we propose to adapt the normal encoder transformer reported in RoCNet~\cite{slimani2023rocnet} which enhances the feature vectors by successive \textit{self-attention} and \textit{cross-attention} layers~\cite{vaswani2017attention} to get the final features $\Tilde{\boldsymbol{f}}_i \in \mathbb{R}^{d}$ and $\Tilde{\boldsymbol{h}}_i \in \mathbb{R}^{d}$. As vanilla transformers used in 3D point clouds may omit geometric structure encoding, authors proposed to feed the transformer with transformation-invariant geometric information. To this aim, for each point, the relative orientation of its surface normal and all the other points of the same set is encoded thanks to sinusoidal functions inspired by the positional encoding from~\cite{vaswani2017attention} and~\cite{geotransformer}. Noting $\boldsymbol{n}_i$ the surface normal estimated on the position  $\boldsymbol{x}_i$, the geometric information between each pair of points is embedded in a vector $\boldsymbol{r}_{i,j} \in \mathbb{R}^{d}  $ as follows: 
\small
\begin{align}
\boldsymbol{r}_{i,j}^{2p} &= \sin\left(\frac{\angle(\boldsymbol{n}_i,\boldsymbol{n}_j) }{\sigma \times u^{2p/{d}}}\right), &
\boldsymbol{r}_{i,j}^{2p+1} &= \cos\left(\frac{\angle(\boldsymbol{n}_i,\boldsymbol{n}_j)}{\sigma \times u^{2p/{d}}}\right)
\end{align}
\normalsize
% \end{equation*}
where $p$ is the dimension index in $\boldsymbol{r}_{i,j}$, $\sigma = \frac{15\times \pi }{ 180} $ is a normalization coefficient to limit the sensitivity of normals orientation, and $u$ a constant empirically defined as $10000$. The embedding $\boldsymbol{r}_{i,j} $ is used in the \textit{self-attention} scores computation next to point features vectors, as detailed in~\cite{slimani2023rocnet}.
%
% -----------
\subsubsection{Matching module}\label{subsec.matching}
Once the final features $\Tilde{\boldsymbol{f}}_i \in \mathbb{R}^{d}$ and $\Tilde{\boldsymbol{h}}_i \in \mathbb{R}^{d}$ built, pair-wise correspondences must be estimated between the two point clouds. To this aim, a similarity matrix $\mathbf{S} \in \mathbb{R}^{M\times N}$ is obtained by a dot product between the feature vectors, with: 
\begin{equation} \label{eq.score_matrix}
\mathbf{S}_{i,j} = \left\langle \Tilde{\boldsymbol{f}}_i ,\Tilde{\boldsymbol{h}}_j \right\rangle   ~ \text{with}  ~1\leq i \leq M , ~1\leq  j \leq N
\end{equation}

This last matrix is incrementally optimized by a differentiable transport algorithm~\cite{sinkhorn1967concerning}. Collecting the mutual best scores along each row and each column from the final score matrix $\Tilde{\mathbf{S}} \in \mathbb{R}^{M\times N}$  allows constructing a binary assignment between the point clouds~\cite{mdgat}.
%
% ---------------
\subsubsection{Transformation estimation}\label{subsec.pose}
The final step of the algorithm is the rigid transformation estimation, which can be computed using a simple Singular Values Decomposition (SVD) of the matched points covariance matrix or by a RANSAC. In this paper, to compare  \emph{LoGDesc} performance with the state-of-the-art methods, we use the farthest sampling-guided registration (FSR) module introduced in RoCNet++~\cite{slimani2024rocnet++} as our pose estimator, its pseudocode is detailed in the pseudocode~\ref{pseudocode}. This is mainly motivated by the fact that it offers a good compromise between robustness to noise and computation cost~\cite{slimani2024rocnet++}. Besides this, to highlight that the proposed feature contributes to the accuracy independently of the pose estimator used, we also compute the rigid transformation using FGR~\cite{zhou2016fast} and RANSAC~\cite{ransac1981}, both with their version based on feature matching (\textit{i.e.}, without using the matching module~\ref{subsec.matching}). We conduct this experiment on the challenging \textit{MVP-RG}~\cite{pan2024robust} dataset and report the results on Table~\ref{tab.mvp.data}. 

%
% -------------
\subsection{Proposed Method}\label{sec.logdesc}
% -------------
%
%
\begin{figure*}[!h]
\centerline{\includegraphics[scale = 0.33,  page=1]{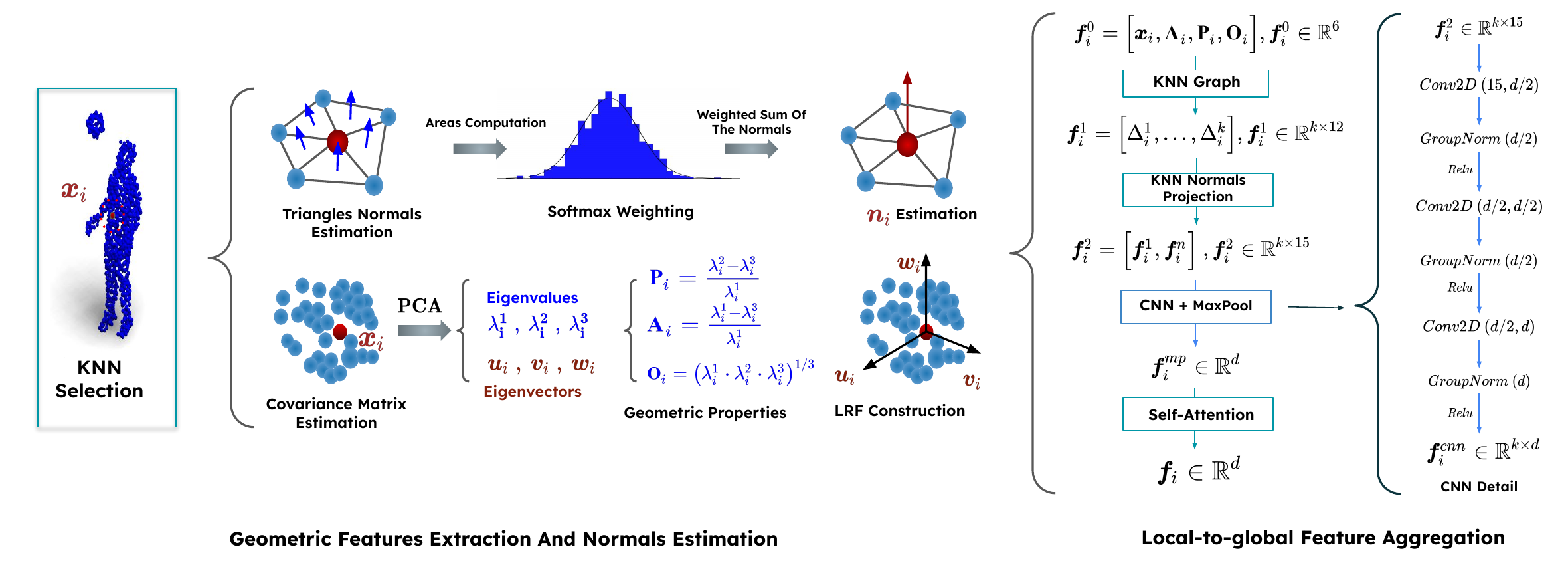}}
\caption{Overview of the proposed geometrical descriptor. {KNN of each point are collected to build local sub-samples. A PCA is applied on these subsamples to extract geometric properties \textbf{P}, \textbf{A} and \textbf{O}, build the {LRF} and compute the normal of each point (\textit{left}). Once the geometric feature vectors built, they are fed into a learning module for convolution and attention features aggregation (\textit{right}) }}
\label{architecture}
\end{figure*}

\scriptsize{
\begin{algorithm}[H]
\SetAlgoLined
\DontPrintSemicolon

\textbf{Input~}{$N$: Number of iterations, $k$: Number of sampled points, }{$\tau$: Inlier threshold, $\boldsymbol{\bar{X}}$, $\boldsymbol{\bar{Y}}$: Paired source and target point clouds}

\textbf{Output~}$\boldsymbol{T}$ \tcp*[r]{\textcolor{blue}{Estimated transformation matrix [4,4]}}

% \smallskip

$\boldsymbol{\hat{T}}$ = \textit{zeros(4,4)} \tcp*[r]{\textcolor{blue}{Initialization with 4x4 null matrix}}
$\boldsymbol{\hat{T}}[3, 3] \gets 1.0$\tcp*[r]{\textcolor{blue}{Set the last cell value}}
$best \gets 0 $\tcp*[r]{\textcolor{blue}{Initialize the best number of inliers}}

\For{$i = 1$ \KwTo $N$}{
    $ind$ = \textit{FPS}($\boldsymbol{\bar{X}}$, $k$) \tcp*[r]{\textcolor{blue}{[$k,1]$, Farthest points indices}}
    $\boldsymbol{\Tilde{X}}$ = $\boldsymbol{\bar{X}}[ind,:]$ \tcp*[r]{\textcolor{blue}{$[k,3]$, Farthest source points gathering}}
    $\boldsymbol{\Tilde{Y}}$ = $\boldsymbol{\bar{Y}}[ind,:]$ \tcp*[r]{\textcolor{blue}{$[k,3]$, Farthest target points gathering}}

    $\boldsymbol{\Tilde{x}_m} = \text{mean}(\boldsymbol{\Tilde{X}}, \text{axis}=0)$ \tcp*[r]{\textcolor{blue}{$[1,3]$, Source points centroid}}
    $\boldsymbol{\Tilde{y}_m} = \text{mean}(\boldsymbol{\Tilde{Y}}, \text{axis}=0)$ \tcp*[r]{\textcolor{blue}{$[1,3]$, Target points centroid}}

    $\boldsymbol{\Tilde{X}}_{\text{c}} \gets \boldsymbol{\Tilde{X}} - \boldsymbol{\Tilde{x}_m}$ \tcp*[r]{\textcolor{blue}{Center source points}}
    $\boldsymbol{\Tilde{Y}}_{\text{c}} \gets \boldsymbol{\Tilde{Y}} - \boldsymbol{\Tilde{y}_m}$ \tcp*[r]{\textcolor{blue}{Center target points}}
    $[\boldsymbol{U},\boldsymbol{S},\boldsymbol{V}] \gets \text{SVD}(\boldsymbol{\Tilde{X}}_{\text{c}}^T \cdot \boldsymbol{\Tilde{Y}}_{\text{c}})$ \tcp*[r]{\textcolor{blue}{SVD of the covariance matrix}}

    $\boldsymbol{R} \gets \boldsymbol{U} \cdot \boldsymbol{V}^T~;~ \boldsymbol{t} \gets \boldsymbol{\Tilde{y}_m} - \boldsymbol{R} \cdot \boldsymbol{\Tilde{x}_m}$ \tcp*[r]{\textcolor{blue}{Compute rotation matrix and translation vector}}
    % $\boldsymbol{t} \gets \boldsymbol{\Tilde{y}_m} - \boldsymbol{R} \cdot \boldsymbol{\Tilde{x}_m}$ \tcp*[r]{\textcolor{blue}{Compute translation vector}}
    
    $\boldsymbol{\hat{T}}[0:3, 0:3] \gets \boldsymbol{R}~;~\boldsymbol{\hat{T}}[0:3, 3] \gets \boldsymbol{t}$ \tcp*[r]{\textcolor{blue}{Set rotation and translation parts}}
    % $\boldsymbol{\hat{T}}[0:3, 3] \gets \boldsymbol{t}$ \tcp*[r]{\textcolor{blue}{Set translation part}}
    $\boldsymbol{\bar{X}}_y \gets \textit{Apply\_transform}(\boldsymbol{\bar{X}}, \boldsymbol{\hat{T}})$ \tcp*[r]{\textcolor{blue}{Transform source points}}
    
    $n_i \gets 0$ \tcp*[r]{\textcolor{blue}{Reset the inlier count for this iteration}}
    \For{$j = 1$ \KwTo $\text{length}(\boldsymbol{\bar{X}}_y)$}{
        $d \gets \| \boldsymbol{\bar{X}}_y[j,:] - \boldsymbol{\bar{Y}}[j,:] \|$ \tcp*[r]{\textcolor{blue}{Pairwise Euclidean distance}}
        \If{$d < \tau$}{
            $n_i \gets n_i + 1$ \tcp*[r]{\textcolor{blue}{Count as inlier if within threshold}}
        }
    }

    \If{$n_i > best$}{
        $best \gets n_i$ \tcp*[r]{\textcolor{blue}{Update best inliers number}}
        $\boldsymbol{T} \gets \boldsymbol{\hat{T}}$ \tcp*[r]{\textcolor{blue}{Update best transformation}}
    }
}

\textbf{Return} $\boldsymbol{T}$ \tcp*[r]{\textcolor{blue}{Best transformation matrix [4,4]}}

\caption{Pseudo-code of the FSR module}
\label{pseudocode}
\end{algorithm}
}
\normalsize

\subsubsection{Geometric features using the 3D covariance matrix}

First, inspired by~\cite{yang2024accurate}, the local structure of each point $\boldsymbol{x}_i$ is described using the three scalar values: the anisotropy  $\textbf{A}_i$, the planarity $\textbf{P}_i$ and the omnivariance $\textbf{O}_i$. These functions are extracted from the 3D covariance matrix $\mathbf{\Sigma}_i$ of the $k$ nearest neighbors of $\boldsymbol{x}_i$ by applying an eigenvalue decomposition on $\mathbf{\Sigma}_i$ to get $\lambda^1_i \geq \lambda^2_i \geq\lambda^3_i $. Then, $\textbf{A}_i$, $\textbf{P}_i$ and $\textbf{O}_i$ are given by:  

\begin{equation}\label{eq.aniso}
 \textbf{A}_i ~ = ~ \frac{\lambda^1_i - \lambda^3_i }{\lambda^1_i}, ~~~ \textbf{P}_i ~ = ~ \frac{\lambda^2_i - \lambda^3_i }{\lambda^1_i}, ~~ \text{and} ~~ \textbf{O}_i ~ = ~ \big( {\lambda^1_i \cdot \lambda^2_i \cdot \lambda^3_i} \big)^{1/3}
\end{equation}

Besides this, the eigenvectors $\boldsymbol{u}_i, \boldsymbol{v}_i$ and $\boldsymbol{w}_i$ associated to  $\lambda^1_i$, $\lambda^2_i$ and $\lambda^3_i$ respectively are used to build a Local Reference Frame (LRF) for each point by stacking them as column vectors in a matrix  $\boldsymbol{R}^{lrf}_i \in \mathbb{R}^{3\times3}$. The latter is used to project the estimated normal vectors detailed in the Section above~(\ref{subsec.normals})
%
% -----------------
\subsubsection{Normal estimation using local triangles}\label{subsec.normals}

Inspired by the umbrella surface representation~\cite{ran2022surface}, we compute the surface normal vector of each point using the $k-1$ triangles formed by its $k$ nearest neighbors. To do so, the normal $\boldsymbol{z}_j$ to the plan containing each triangle $j$ is computed using a simple cross product between its two edges. The final normal vector $\boldsymbol{n}_i$ associated to the point $\boldsymbol{x}_i$ is computed by a weighted sum of the $k-1$ triangle normals: 

\begin{equation}\label{eq.normals_enc}
 \boldsymbol{n}_i = \sum_{j=1}^{k-1} \big( \omega_j \boldsymbol{z}_j \big) 
\end{equation}
where the weights $\omega_j$ are the result of a \textit{SoftMax} operation over the areas of each triangle to give a bigger contribution for the largest triangles and reduce the smallest triangles contribution since experiments showed they are more sensitive to the presence of noise.
%
% -------------
\subsubsection{Local-to-global feature aggregation}\label{subsec.feature_conc}
Once the geometrical properties $\textbf{A}_i$, $\textbf{P}_i$ and $\textbf{O}_i$ and the normal $ \boldsymbol{n}_i$ are estimated for each single point, we propose to construct an initial descriptor vector $\boldsymbol{f}^0_i$ by concatenating the 3D coordinates of each  $\boldsymbol{x}_i$  and its associated $\textbf{A}_i$, $\textbf{P}_i$ and $\textbf{O}_i$: 
\begin{equation}\label{eq.feat0}
 \boldsymbol{f}^0_i = \Big[ \boldsymbol{x}_i, \textbf{A}_i, \textbf{P}_i, \textbf{O}_i  \Big]~, \boldsymbol{f}^0_i \in \mathbb{R}^6
\end{equation}

Inspired by~\cite{dgcnn}, we propose to propagate this local information by comparing it to the feature vectors of the $k$ nearest neighbors of $\boldsymbol{x}_i$ to get a new intermediate geometrical descriptor: 
\begin{equation}\label{eq.feat1}
 \boldsymbol{f}^1_i = \Big[ \Delta^1_i,...,\Delta^k_i  \Big] ~, \boldsymbol{f}^1_i \in \mathbb{R}^{k\times12}
\end{equation}
where $\Delta^j_i$ is the vector capturing the information between $\boldsymbol{x}_i$, and its $j^{th}$ neighbour. It is obtained by the following: 
\begin{equation}\label{eq.delta}
 \Delta^j_i = \big[ \boldsymbol{f}^0_i , (\boldsymbol{f}^0_j-\boldsymbol{f}^0_j) ]  ~,  \Delta^j_i \in \mathbb{R}^{1\times12}
\end{equation}

To better exploit the very local structure around each point, the normal vectors $\boldsymbol{n}_i$ complete the last vector and get the final geometrical descriptor $ \boldsymbol{f}^2_i $. To do so, we propose to collect the normals of the same $knn$ and project them in the $LRF$ estimated previously. The projected normal of the $j^{th}$ neighbour noted $ \boldsymbol{n}^{lrf}_j$ is thus obtained  by the following:  
\begin{equation}\label{eq.proj_lrf}
 \boldsymbol{n}^{lrf}_j = \boldsymbol{R}^{lrf}_i \cdot \boldsymbol{n}_j 
\end{equation}

The projected vectors of the $knn$ are then stacked, resulting in a new vector $  \boldsymbol{f}^n_i \in \mathbb{R}^{k\times3} $. The final descriptor $\boldsymbol{f}^2_i $ is the concatenation of  $ \boldsymbol{f}^n_i$ and  $ \boldsymbol{f}^1_i$: 
\begin{equation}\label{eq.f2}
\boldsymbol{f}^2_i  =  \Big[  \boldsymbol{f}^1_i , \boldsymbol{f}^n_i  \Big] ~, \boldsymbol{f}^2_i  \in \mathbb{R}^{k\times15}
\end{equation}

The so-obtained descriptor encodes the variation of anisotropy, planarity, and omnivariance between the point and its neighbors, as well as the variation of the normals. The same procedure is repeated for each point from the source and the target point clouds. To improve the robustness of the points representation, the features are fed to successive learning modules. {Inside each point cloud, all the points and their knn form a graph feature tensor $\mathrm{R}^{N\times k\times15}$ which is fed into a CNN composed of three successive 2D convolutions with $1\times1$ kernel size to increase the features dimension from $15$ to $d$ resulting in a new feature tensor $\mathrm{R}^{N\times k\times d}$}. Each layer is followed by a group normalisation operation and a \textit{ReLU} activation function to get updated features $\boldsymbol{f}^{cnn}_i \in \mathbb{R}^{k\times d} $ for each point. A \textit{MaxPool} operation over the $k$ neighbours is applied to obtain a single dimensional features vector $\boldsymbol{f}^{mp}_i = MaxPool(\boldsymbol{f}^{cnn}_i) \in \mathbb{R}^{d}$. Finally, to propagate a global intra-point cloud representation for the two input sets, a \textit{self-attention} mechanism is introduced. It contains three layers incorporating the 3D rotatory position encoding from~\cite{li2022lepard} and~\cite{su2024roformer} to build an efficient and translation-invariant representation within the \textit{self-attention} module. This is achieved by rotating the descriptor of each point with a matrix $\boldsymbol{R}^{sa} \in \mathbb{R}^{d\times d}$ defined as detailed below: 
\begin{equation*}\label{eq.Rot_sa}
\boldsymbol{R}^{sa}  = Diag \big[ \boldsymbol{R}^1 , \boldsymbol{R}^2,\dots, \boldsymbol{R}^{d/6}   \big]
\end{equation*}
where each $\boldsymbol{R}^{j}$ is a $\mathbb{R}^{6\times 6} $ matrix defined as follows:
\scriptsize
\begin{equation}\label{eq.block_matrix}
\footnotesize
\boldsymbol{R}^j  = 
\resizebox{0.8\linewidth}{!}{
\(
\begin{bmatrix}
   \cos{x\theta_j}  & -\sin{x\theta_j} &  0 & 0 & 0 & 0 \\
   \sin{x\theta_j}  &  \cos{x\theta_j} &  0 & 0 & 0 & 0 \\
   0 & 0 & \cos{y\theta_j}  & -\sin{y\theta_j} &  0 & 0  \\
   0 & 0 & \sin{y\theta_j}  &  \cos{y\theta_j} &  0 & 0  \\
   0 & 0 & 0  & 0 & \cos{z\theta_j}  & -\sin{z\theta_j}   \\
   0 & 0 & 0  & 0 & \sin{z\theta_j}  &  \cos{z\theta_j}  \\
\end{bmatrix}
\)
}
\end{equation}
\normalsize
where $x$, $y$ and $z$ are the spatial coordinates of the point $\boldsymbol{x}_i$ and $\theta_j$ is the embedding associated to the $j^{th}$ dimension given by $\theta_j = 1/\big({10000^{6(j-1)/d}}\big)$. The final feature vector is then updated thanks to the following equations: 
\begin{equation}\label{eq.attention_vec}
\boldsymbol{f}^{mp}_i  \leftarrow  \boldsymbol{f}^{mp}_i + MLP  \big[ \boldsymbol{R}^{sa}\mathbf{W}^Q\boldsymbol{f}^{mp}_i , \sum_{j=1}^{M} \alpha_{ij} \big( \boldsymbol{R}^{sa}\mathbf{W}^V\boldsymbol{f}^{mp}_j \big)  \big] 
\end{equation}
with $[\cdot,\cdot]$ is the concatenation operation, $j$ covering all the points contained in the source point cloud   $\boldsymbol{X} \in \mathbb{R}^{M\times3}$, and: 
\begin{equation}\label{eq.rotated_vec}
\alpha_{ij} = \underset{j}{softmax} \bigg( \frac{(\boldsymbol{R}^{sa}\mathbf{W}^Q\boldsymbol{f}^{mp}_i )(\boldsymbol{R}^{sa}\mathbf{W}^K\boldsymbol{f}^{mp}_j )^T}{\sqrt{{d}}}\bigg)
\end{equation}
where $\mathbf{W}^Q \in \mathbb{R}^{d\times d}$, $\mathbf{W}^K \in \mathbb{R}^{d\times d} $ and $\mathbf{W}^V \in \mathbb{R}^{d\times d} $ are the learned projection matrices for the query, the key and the value points in the attention message passing. The exact same operations are applied for all the points $\boldsymbol{y}_i$ from the target point cloud   $\boldsymbol{Y}  \in \mathbb{R}^{N\times3}$. The final descriptor of each source point $\boldsymbol{x}_i \in \boldsymbol{X} $ will be referred to as $\boldsymbol{f}_i \in \mathbb{R}^{d}$. At the same time, we call $\boldsymbol{h}_i \in \mathbb{R}^{d}$, the final descriptor of each target point $\boldsymbol{y}_i \in \boldsymbol{Y} $. We then follow the algorithm described in  Section~\ref{sec.previous_work} for point matching, and we adopt the Farthest Sampling-guided Registration~\cite{slimani2024rocnet++} for the rigid transformation estimation. 
%
% ---------------------
\section{EXPERIMENTS}\label{sec.eval}
% ---------------------
%
% ---------------
%
The method is implemented in PyTorch and trained on a Nvidia Tesla V100-32G GPU using the Adam optimiser~\cite{kingma2014adam} with a learning rate of $10^{-4}$ for 100 epochs for the \textit{ModelNet40} dataset and 200 epochs for the \textit{MVP-RG} dataset. A number of $k = 30$ nearest neighbors are used to construct the local triangles and graphs, while the feature dimension $d$ is set to $132$. The kernel size of all the convolutions is set to $1\times1$, and the \textit{GroupNorm} operations are followed by a \textit{ReLU} activation function. The output features of the CNN $\boldsymbol{f}^{mp}_i \in \mathbb{R}^{132}$ are fed into $4$ self-attention layers. For the LRF and the geometric features $\textbf{A}$, $\textbf{P}$ and $\textbf{O}$ estimation, we collect a maximum of $128$ neighbors and keep only those that are less than $r = 0.3$ away from the point on which the PCA is applied.
%
% ---------------
\subsection{Datasets}\label{sec.datasets}
To evaluate \emph{LoGDesc}  performances, we first test it on the synthetic \textit{ModelNet40} dataset~\cite{wu20153d}. This last proposes 9,843 point clouds for training and 2,468 for testing, containing $2048$ points sampled from CAD models. As in~\cite{wang2019prnet}, the target point clouds are created by randomly rotating each set by an angle between $0$ and $45$ degrees and translating them by a displacement between $-0.5$ and $0.5$ along each axis. This is followed by a random permutation of the points before  $1024$ points are selected for each object as input. Besides this, we follow~\cite{wang2019prnet} and test the robustness to noise and to outliers of \emph{LoGDesc}, first, by adding a Gaussian noise sampled from $N$ (0, 0.01) and clipped to [0.05, 0.05] to each point. Finally, occlusions are simulated by selecting the $768$ nearest neighbors from the source and the target point clouds of a random point in space. For a second time, we propose to follow~\cite{r_pointhop} and assess the ability of \emph{LoGDesc} to generalize to unseen objects and real point clouds by testing the model trained on ModelNet using the $10$ range scanner point clouds from \textit{Stanford Bunny} dataset~\cite{turk1994zippered}. The root mean squared error (RMSE) and mean absolute error (MAE) between ground truth values and estimated values for rotation and translation are reported in the following section for registration evaluation. At the same time, precision (P), accuracy (A), and recall (R) are used to analyze the estimation of point correspondence. Finally, we assess our method on MVP (Multi-View Partial)-RG dataset~\cite{pan2024robust}, which contains $7600$ partial point clouds representing $16$ different objects categories generated by virtual cameras from diverse viewpoints leading to inconsistent local point densities and thus making this dataset extremely challenging. We perform the exact same study than~\cite{pan2024robust} by splitting the dataset into $6400$ training samples and $1200$ testing samples and using the same metrics: isotropic rotation errors $L_R$, translation errors $L_t$, and the root mean square error $L_{RMSE}$.  {We propose to recreate the USIP}~\cite{li2019usip} and MDGAT~\cite{mdgat} {experiments on \textit{KITTI} where $256$ keypoints are used for registration to address the computational cost of our method on large point clouds, as noted in the paper's conclusion. We follow the same training, testing procedures, and metrics as MDGAT (Failure Rate \textbf{FR} and the Inlier Ratio \textbf{IR} ).}

Finally, we aim to explore another potential application of \emph{LoGDesc}, i.e., medicine as outlined by~\cite{hu2021markerless} dealing with knee arthroplasty surgery. Specifically, we plan to evaluate the model trained on \textit{ModelNet40} using a set of point clouds representing a human femur. This dataset includes point clouds captured with a 12 MPx TrueDepth smartphone camera and a Kinect containing approximately $10,000$ points. Additionally, we will randomly sample $5,000$ points from the CAD model to generate an input point cloud for the model. For evaluation, $2,048$ points will be randomly sampled from the initial point clouds, following the same procedure as \textit{ModelNet40} to simulate partial overlap and noisy input.
% -------------------
\subsection{Results} 
%-------------------
In the matching challenge, Table~\ref{tab.matching} shows that \emph{LoGDesc}  outperforms the state-the-art descriptors DGCNN~\cite{dgcnn}, FPFH~\cite{rusu2009fast} and the triangle-based descriptor~\cite{slimani2024rocnet++} in all the metrics on the noisy point clouds, thus demonstrating the robustness of the proposed method to noise. Concerning the transformation estimation, experiments show that the good results in the matching challenge highlighted previously are reflected in the estimation of the rigid transformation as shown in Tables~\ref{tab.full.noisy.data}  and ~\ref{tab.partial.noisy.data}. Indeed, \emph{LoGDesc} outperforms all tested methods in three of the four metrics when applied to occlusion-free point clouds, is second in translation RMSE behind R-PointHop~\cite{r_pointhop}, and is first in all metrics on partially overlapping point clouds, ex-\oe quo with RoCNet++ in translation, both methods outperforming the second best (RoCNet~\cite{slimani2023rocnet} and GeoTransformer~\cite{geotransformer}) by $50\%$ in RMSE and by $67\%$ in MAE. {On this challenging case of noisy partially overlapping point clouds, we also report in Table}~\ref{tab.modelnet.rigametrics} { the results under the relative rotation  $L_{R}$ and relative translation   $L_{t}$  errors  and  the RMSE  on the rigid transformation $L_{RMSE}$ metrics, as proposed in}~\cite{yu2024riga}. Table~\ref{tab.modelnet.rigametrics}{ confirms the strong performances seen in the previous table, since it matches the best performance in $L_{t}$ and  $L_{RMSE}$ while ranking second in $L_{R}$}. Our method shows an interesting generalization ability since the model trained on \textit{ModelNet40} achieved, on the \textit{Stanford Bunny} dataset, the best result in the RMSE($\mathbf{t}$), the second in the MAE($\mathbf{R}$) while being third in RMSE($\mathbf{R}$) right behind RoCNet (with a $2\% $ gap) and fourth in the MAE($\mathbf{t}$) as shown in Table~\ref{tab.bunny_metrics}. 
\begin{table}%[!h]
\centering
\begin{center}
    
\caption{Matching performances on \textit{ModelNet40} with noisy and partially overlapping point clouds}
\begin{tabularx}{\linewidth}{X | X  X  X}
\hline 

\hline 

\hline

\multicolumn{1}{l|}{{Descriptor}} & \textbf{P}($\uparrow$)  & \textbf{A}($\uparrow$)  & \textbf{R}($\uparrow$)   \\
\hline

\multicolumn{1}{l|}{{FPFH}~\cite{fpfh}}  & 72.1 & 71.0 & 71.2  \\
\multicolumn{1}{l|}{{DGCNN}~\cite{dgcnn}}  & {85.8} & {85.9} & {85.5}  \\
\multicolumn{1}{l|}{{{RoCNet++}~\cite{slimani2024rocnet++}}}  & \underline{89.4} & \underline{89.4} & \underline{89.2}  \\

\hline

{\textbf{Ours}} &\textbf{92.0} & \textbf{92.1} & \textbf{91.9} \\
\hline 

\hline 

\hline
\end{tabularx}
\label{tab.matching}

\end{center}

% %\end{adjustbox}
\end{table}

Moreover, Table~\ref{tab.mvp.data} highlights that our method can handle very challenging configurations with varying local densities, low overlapping point clouds, and unrestricted rotations $[0^\circ,360^\circ]$ since it outperforms the other methods in all the used metrics by a $55\%$ relative gap at least. The last two rows of the same Table also highlight that \emph{LoGDesc} can perform well when combined with other pose estimators than FSR since the registration performances still make our method the best in three metrics when the matching process detailed in Section~\ref{subsec.matching} is skipped. The transformation is estimated using RANSAC or FGR based on feature matching. {Table}~\ref{tab.kitti} {indicates strong abilities of \emph{LoGDesc} to handle low overlapping and real world sources data (\textit{i.e} Lidar) which may be corrupted with sensor noise, since it comes first for \textbf{FR} and second for \textbf{IR} on the \textit{KITTI} keypoints registration. }

\begin{figure}[!h]
\centerline{\includegraphics[width=\columnwidth]{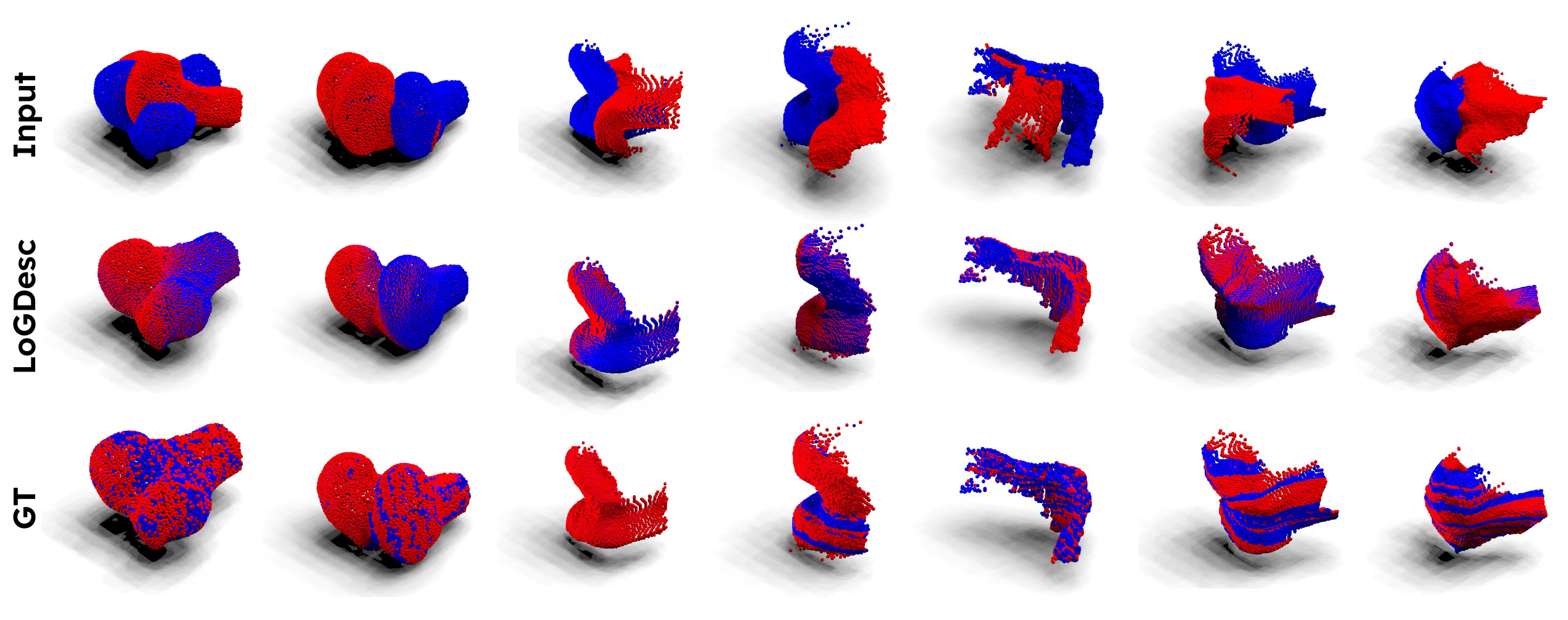}}
\caption{Performed registrations on a 3D printed femur.}
\label{fig.femur}
\end{figure}

An additional experiment was carried out using real data from a human femur as described in section~\ref{sec.datasets}. Figure~\ref{fig.femur} shows a sample of the registration results. The results indicate that \emph{LoGDesc} can effectively generalise to unseen and real data, provided that the desired transformation matrix is within the range seen during training. The performance of \textit{LoGDesc} is promising, especially given its ability to generalise to other data, especially in certain applications where annotated data is difficult to obtain, such as medical applications like neurosurgery and orthopaedics.
\begin{table}%[!h]
\caption{Performances on noisy and fully overlapping data from \textit{ModelNet40}.}

\begin{center}
\begin{tabular}{l | l  l | l l}
\hline 

\hline 

\hline
Method & RMSE($\mathbf{R}$) & MAE($\mathbf{R}$) & RMSE($\mathbf{t}$) & MAE($\mathbf{t}$)  \\ 
\hline

%ICP~\cite{besl1992method}  & 11.971         & 4.497         & 0.04832       & 0.00433 \\
DCP-V2~\cite{dcp}    & 8.417         & 5.685         & 0.03183       & 0.02337    \\
PRNET~\cite{wang2019prnet}     & 3.218         & 1.446         & 0.11178       & 0.00837  \\
R-PointHop~\cite{r_pointhop} & 2.780         & 0.980         & \textbf{0.00087}  &  {0.00375} \\
VRNet~\cite{zhang2022vrnet}  & {2.558} & {1.016} & {0.00570} & 0.00289  \\
GeoTransf~\cite{geotransformer}       & \underline{0.692}    &  {0.267} &  0.00519  & 0.00200  \\
{RoCNet}~\cite{slimani2023rocnet}&  {1.920}  & {0.555} & {0.00260} & {0.00180}  \\
{RoCNet++}~\cite{slimani2024rocnet++} & {{1.004}}  &  \underline{0.249} &  {0.00133}  & \underline{0.00092} \\

\hline 

{\textbf{Ours}}&  \textbf{0.618}  &  \textbf{0.187} & \underline{0.00121} & \textbf{0.00089} \\ 

\hline 

\hline 

\hline
\end{tabular}
\label{tab.full.noisy.data}
\end{center}
\end{table}
\normalsize
%
% -----------------------------------
 
\begin{table}%[!h]
\caption{Performances on noisy and partially overlapping data from \textit{ModelNet40}.}

\begin{center}
\begin{tabular}{l | l  l | l l}
\hline 

\hline 

\hline
Method & RMSE($\mathbf{R}$) & MAE($\mathbf{R}$) & RMSE($\mathbf{t}$) & MAE($\mathbf{t}$)  \\ 
\hline

DCP-V2~\cite{dcp} & 6.883 & 4.534 & 0.028 & 0.021 \\
PRNET~\cite{wang2019prnet}& 4.323  & 2.051 & 0.017 & 0.012 \\
VRNet~\cite{zhang2022vrnet}& 3.615 & 1.637 & 0.010 & 0.006 \\
GeoTransf~\cite{geotransformer} & \underline{0.915}  &  {0.386}     & {0.007} & \underline{0.003}\\
{RoCNet}~\cite{slimani2023rocnet} & {1.810}  &{0.620} & \underline{0.004} & \underline{0.003} \\
{{RoCNet++}~\cite{slimani2024rocnet++}} &  {1.278} & \underline{0.318} & \textbf{0.002} & \textbf{0.001} \\
\hline 

{\textbf{Ours}}  & \textbf{0.774}  & \textbf{0.266} & \textbf{0.002} & \textbf{0.001} \\

\hline 

\hline 

\hline
\end{tabular}
\label{tab.partial.noisy.data}
\end{center}
\end{table}

\begin{table}%[!h]
% \begin{highlighttcb}

\caption{{Performances on noisy and partial data from \textit{ModelNet40} using the metrics from}~\cite{yu2024riga}} 
\centering
\begin{tabular}{l | l  |  l | l | l | l | l }
\hline

\hline

\hline

& PRNet~\cite{wang2019prnet} & DCP~\cite{dcp} & Predator~\cite{huang2021predator} & GMCNet~\cite{pan2024robust} & RIGA~\cite{yu2024riga} & \textbf{Ours} \\
\hline
$L_{R}$ & 4.37$^\circ$ & 9.33$^\circ$ & 3.33$^\circ$ & \textbf{0.94$^\circ$} & 1.15$^\circ$ & \underline{0.95$^\circ$} \\
$L_{t}$ & 0.034 & 0.070 & 0.018 & 0.007 & \textbf{0.006} & \textbf{0.006} \\
$L_{RMSE}$ & 0.045 & 0.018 & 0.025 & \textbf{0.008} & \underline{0.009} & \textbf{0.008} \\

\hline

\hline

\end{tabular}

\label{tab.modelnet.rigametrics}
% \end{highlighttcb}

\end{table}

%\normalsize

\begin{table}%[!h]

\caption{Performances on the \textit{Bunny Stanford }dataset}

\begin{center}

\begin{tabular}{l | l  l | l l}
\hline

\hline 
 
\hline
Method & RMSE($\mathbf{R}$) & MAE($\mathbf{R}$) & RMSE($\mathbf{t}$) & MAE($\mathbf{t}$)  \\ 
\hline

%ICP~\cite{besl1992method}  & 13.32 & 10.72 &  0.0492 & 0.0242 \\
%Go-ICP  & 12.92 & 4.52 &  0.0429 & 0.0282 \\
FGR~\cite{zhou2016fast} & 1.99 & 1.49  & 0.1993 & 0.1658 \\
DCP~\cite{dcp}  & 6.44 & 4.78  & 0.0406 & 0.0374 \\
R-PointHop~\cite{r_pointhop}  & 1.49 & 1.09  & 0.0361 & \textbf{0.0269} \\
{RoCNet}~\cite{slimani2023rocnet} &  \underline{0.99} & {0.83}  & {0.0338} & \underline{0.0288} \\
{RoCNet++}~\cite{slimani2024rocnet++} & \textbf{0.91} & \textbf{0.50} & \underline{0.0316} & {0.0301} \\

\hline

{\textbf{Ours}}  & 1.01 & \underline{0.73} & \textbf{0.0272} & {0.0310} \\

\hline

\hline 
 
\hline

\end{tabular}
%\end{adjustbox}
\label{tab.bunny_metrics}
\end{center}
\end{table}
    
%\small

\begin{table}%[!h]

\caption{{Performances on Multi-View Partial virtual scan ReGistration (\textit{MVP-RG}). The indication $^{(\dagger)}$ indicates a pose estimation based on  \emph{LoGDesc} features matching, without the matching module~\ref{subsec.matching}}} 
\centering
\begin{tabular}{l | l  |  l | l }
\hline

\hline

\hline

Method & $L_{R}$ & $L_{t}$ & $L_{RMSE}$\\
\hline
IDAM~\cite{li2020iterative}    & 24.35$^\circ$ & 0.280 & 0.344 \\
RGM~\cite{rgm}     & 41.27$^\circ$ & 0.425 & 0.583 \\
DCP~\cite{dcp}    & 30.37$^\circ$ & 0.273 & 0.634 \\
% DeepGMR & 43.74$^\circ$ & 0.353 & 0.608 \\
{Predator} \cite{huang2021predator} & \underline{10.58}$^\circ$ & \underline{0.067} & \underline{0.125}  \\
RPMNet~\cite{yew2020rpm}  & 22.20$^\circ$ &  {0.174} & 0.327 \\
GMCNet~\cite{pan2024robust}  & {16.57$^\circ$} & {0.174} & {0.246} \\
\hline

{\textbf{Ours}}& \underline{7.33$^\circ$}  &  \textbf{0.043} &  \underline{0.099}  \\

{\textbf{Ours$^{(\dagger)}$ - RANSAC (1k)}}& \textbf{6.64$^\circ$}  &  \underline{0.053} &  \textbf{0.082}  \\ 

{\textbf{Ours$^{(\dagger)}$ - FGR}}& {9.08$^\circ$}  &  {0.061} &  \underline{0.099}  \\ 

\hline

\hline

\end{tabular}

\label{tab.mvp.data}

\end{table}
%\normalsize

\begin{table}%[!h]
\centering
\caption{ {Registration performance on \textit{KITTI} using $256$ USIP keypoints.}}
\begin{tabular}{l|c|c|c|c|c|c|c|c}
\hline
\textbf{Metric} & \makecell{\textbf{FPFH}\\~\cite{fpfh}} & \makecell{\textbf{SHOT}\\~\cite{shi2020points}} & \makecell{\textbf{3DFeatNet}\\~\cite{yew2018-3dfeatnet}} & \makecell{\textbf{USIP}\\~\cite{li2019usip}} & \makecell{\textbf{DCP}\\~\cite{dcp}} & \makecell{\textbf{MDGAT}+\\\textbf{SuperGlue}\\~\cite{mdgat,sarlin20superglue}} & \makecell{\textbf{MDGAT}\\~\cite{mdgat}} & \textbf{Ours} \\
\hline
\textbf{FR} & 8.37 & 5.40 & 1.55 & 1.41 & 4.01 & \underline{0.58} & 0.67 & \textbf{0.52} \\
\textbf{IR} & 18.77 & 18.21 & 22.48 & 32.20 & 35.37 & 36.19 & \textbf{42.23} & \underline{37.7} \\
\hline
\end{tabular}
\label{tab.kitti}
\end{table}

%% Big figure

\begin{figure}%[!h]
\centerline{\includegraphics[width=\columnwidth]{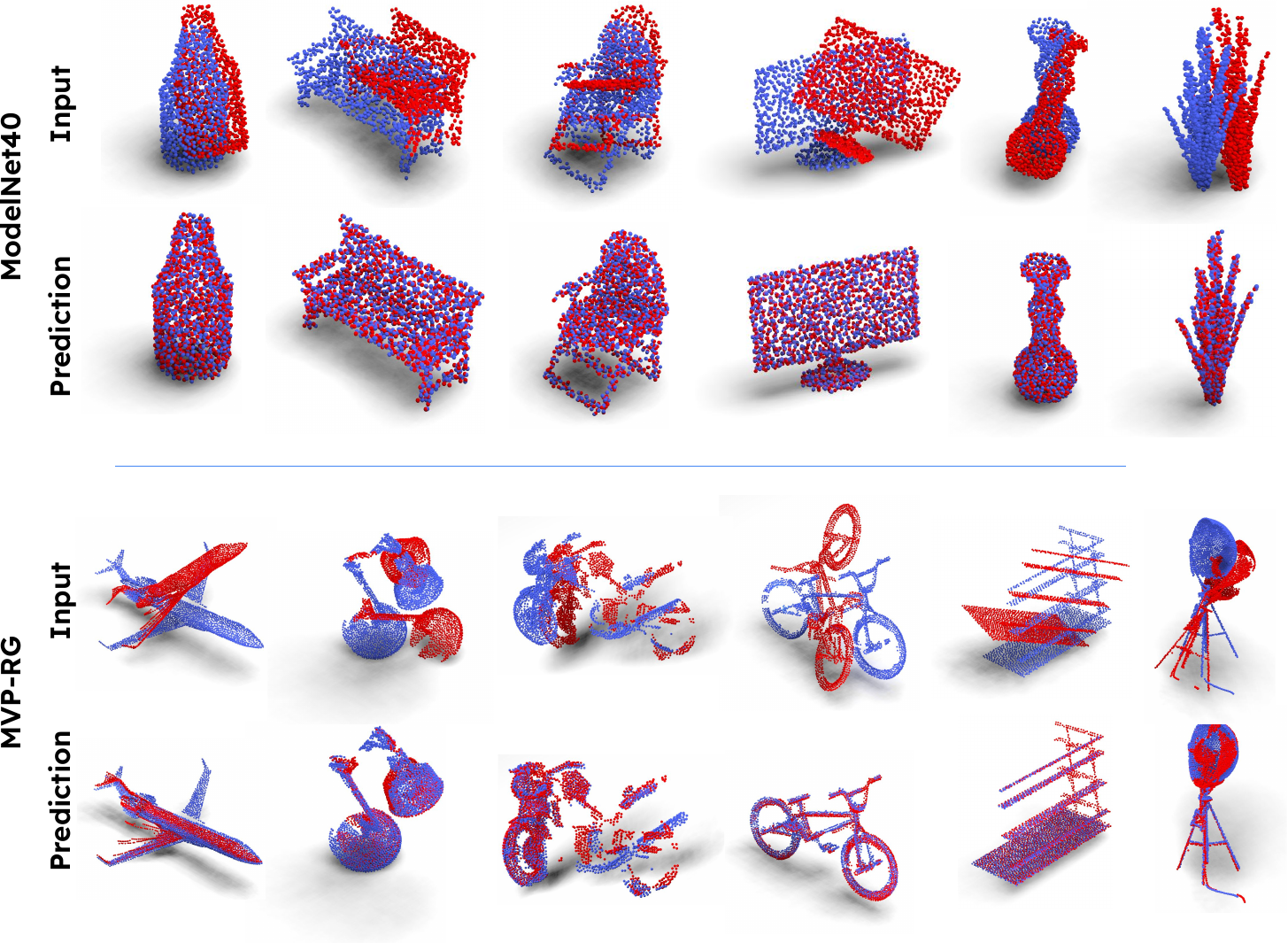}}
\caption{Performed registrations on \textit{ModelNet40} (top) and \textit{MVP-RG} (bottom).}
\label{fig.examples}
\end{figure}

%
% -------------------
\subsection{Ablation Study} 

{To highlight the impact of using the geometric properties proposed in LoGDesc, an ablation study is proposed here under noisy, partially overlapping point clouds from \textit{ModelNet40}. In} Table~\ref{tab.ablation},{ different versions of \emph{LoGDesc} are tested by removing each geometric variable -\textbf{A},\textbf{P}, \textbf{O} and the normals (\textbf{N}) - one by one, leaving the rest of the architecture unchanged. The results show that each variable has its contribution to the descriptor performance, with in particular the removal of \textbf{A}, \textbf{O}, and \textbf{P} leading to a 7\% drop in matching metrics.}

\begin{table}[!h]
\centering
\caption{{ Matching performances on noisy data (\textit{ModelNet40})}}
\begin{tabular}{l | c | c | c | c | c | c }
\hline
\multicolumn{1}{l|}{\textbf{Metric}} & w/o  \textbf{A,O,P} & w/o  \textbf{A} & w/o  \textbf{O} & w/o  \textbf{P} & w/o  \textbf{N} & \textbf{Ours} \\
\hline
\textbf{P}($\uparrow$) & 85.2 & 88.7 & 91.6 & 88.8 & 91.6 & \textbf{92.0} \\
\textbf{A}($\uparrow$) & 85.4 & 88.8 & 91.7 & 89.0 & 91.7 & \textbf{92.1} \\
\textbf{R}($\uparrow$) & 85.0 & 88.4 & 91.5 & 88.7 & 91.5 & \textbf{91.9} \\
\hline
\end{tabular}

\label{tab.ablation}
\end{table}

%
% ---------------------
\section{CONCLUSION}\label{sec.conclusion}
% ---------------------
%
This paper presented a new robust descriptor called \emph{LoGDesc} and a deep learning architecture for point cloud registration. This descriptor exploits local and global geometric properties of 3D points and machine learning techniques for robust feature extraction and point matching. By aggregating local geometric information such as flatness, anisotropy, and omnivariance, as well as features learned through graph convolutions and attention mechanisms,  \emph{LoGDesc} demonstrates superior performance, especially in handling noisy point clouds and challenging registration scenarios where most methods in the literature show limitations in terms of robustness and precision. An evaluation of our method was carried out on the \textit{ModelNet40}, \textit{Stanford Bunny}, \textit{MVP-RG} and \textit{KITTI} datasets, highlighting the effectiveness and robustness of  \emph{LoGDesc}, which surpasses the most advanced and recent methods in the literature, especially on noisy data. 

For future work, we plan to extend our method to other robotics tasks such as object recognition, visual servoing, and 6 DoF multi-object pose estimation. We will also focus on improving the scalability of  \emph{LoGDesc} to handle larger cloud points, which currently pose a computational problem due to the use of attention mechanisms in the pipeline using, for instance, the superpoints concept. 

\begin{credits}
\subsubsection{\ackname}This work was supported by the French ANR program MARSurg (ANR-21-CE19-0026).

\subsubsection{\discintname}
 The authors have no competing interests to declare that are
relevant to the content of this article.
\end{credits}

\bibliographystyle{splncs04}
\bibliography{main}
\end{document}